\documentclass{article}

\usepackage{arxiv}

\usepackage[utf8]{inputenc} 
\usepackage[T1]{fontenc}    
\usepackage{hyperref}       
\usepackage{url}            
\usepackage{booktabs}       
\usepackage{amsfonts}       
\usepackage{nicefrac}       
\usepackage{microtype}      
\usepackage{lipsum}		
\usepackage{graphicx}
\usepackage{natbib}
\usepackage{doi}
\usepackage{verbatim}
\usepackage{subcaption}
\usepackage[misc]{ifsym}

\newcommand{\behdescription}[0]{
Note: for feature labels refer to Table 1, AQ - \textit{Asking questions}, GQ - \textit{Giving questions, to students...}, O - \textit{Organization: giving class outline...}, S - \textit{Session on tests}, AT - \textit{Active teacher stands by slides...}, SU - \textit{Summing up}
}

\title{A Deep Learning Approach for Automatic Detection of Qualitative Features of Lecturing}

\author{ 
 \href{https://orcid.org/0000-0002-3407-7570}{\includegraphics[scale=0.06]{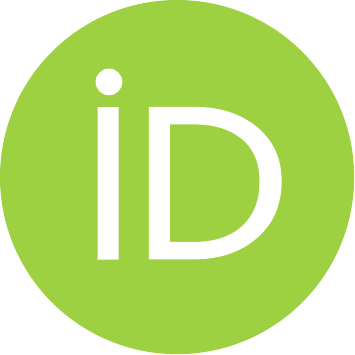}\hspace{1mm}Anna Wr{\'o}blewska$^1${\Letter}}
 	\And J{\'o}zef Jasek$^1$ 
 	\And Bogdan Jastrz{\c e}bski$^1$ 
 	\And 
\href{https://orcid.org/0000-0001-7511-9995}{\includegraphics[scale=0.06]{orcid.pdf}\hspace{1mm}Stanis{\l}aw Pawlak$^1$}
 	\And Anna Grzywacz$^1$ \\
 	\And 
\href{https://orcid.org/0000-0002-8589-6699}{\includegraphics[scale=0.06]{orcid.pdf}Cheong Siew Ann$^2$} 
 	\And 
\href{https://orcid.org/0000-0002-8406-0144}{\includegraphics[scale=0.06]{orcid.pdf}Tan Seng Chee$^2$}
 	\And 
\href{https://orcid.org/0000-0002-1486-8906}{\includegraphics[scale=0.06]{orcid.pdf}Tomasz Trzci{\'n}ski$^{1,3,4}$} 
 	\And 
\href{https://orcid.org/0000-0003-2645-0037}{\includegraphics[scale=0.06]{orcid.pdf}Janusz Ho{\l}yst$^1$} \\
 	$^1$Warsaw University of Technology, $^2$Nanyang Technological University, \\$^3$Jagiellonian University of Cracow, $^4$Tooploox \\
 	\Letter~A.W. -- \texttt{anna.wroblewska1@pw.edu.pl}
 }

\date{}

\renewcommand{\shorttitle}{Automatic Detection of Qualitative Features of Lecturing}

\hypersetup{
pdftitle={A Deep Learning Approach for Automatic Detection of Qualitative Features of Lecturing},
pdfsubject={cs.CY, cs.MM, cs.AI, cs.LG},
pdfauthor={Anna Wróblewska, Józef Jasek, Bogdan Jastrzębski, Stanisław Pawlak, Anna Grzywacz, Cheong Siew Ann, Tan Seng Chee, Tomasz Trzciński, Janusz Hołyst},
pdfkeywords={Didactic features, Qualitative analysis, Video recognition, Deep learning},
}

\begin{document}
\maketitle

\begin{abstract}
	Artificial Intelligence in higher education opens new possibilities for improving the lecturing process, such as enriching didactic materials, helping in assessing students' works or even providing directions to the teachers on how to enhance the lectures. We follow this research path, and in this work, we explore how an academic lecture can be assessed automatically by quantitative features. First, we prepare a set of qualitative features based on teaching practices and then annotate the dataset of academic lecture videos collected for this purpose. We then show how these features could be detected automatically using machine learning and computer vision techniques. Our results show the potential usefulness of our work. 
\end{abstract}


\section{Introduction}
Digital lecturing proliferation during the pandemic drew public attention to the higher education situation and its possible improvements~\cite{crawford2020covid}. Numerous newly created programs aiming to enhance education quality use the advantage of artificial intelligence and deep learning methods~\cite{doi:10.1111/bjet.13018}. 
Adopting artificial intelligence (AI) in higher education is still a significant challenge. The principal aim of this research is to improve the academic lecturing process. When the majority of newly proposed solutions focus on the learners~\cite{kuvcak2018machine}, their performance~\cite{anand2018recursive}, or learning sources~\cite{zhu2015machine}, this research concentrates on the teachers. 

The lecturing process requires very high academic specialization of the lecturers and their high didactic abilities. Our research explores how an academic lecture can be assessed automatically by quantitative features. We define the features following teaching practices and their crucial aspects. The features also regard such techniques as a slide presentation and other visualizations, organizational information, and teacher interactions with the public. The main goal of designing these features is to give objective feedback to the lecturers to improve their didactic behaviours or the content of their lecture materials. On the other hand, we designed the features to be feasible to detect automatically using artificial intelligence methods, especially computer vision or machine learning.

After designing features, we collect and annotate a dataset with those features. Then we tested how we need to preprocess and prepare deep learning models to detect these features. Our general goal is to build a solution that can automatically help teachers get feedback about their lecture assessment and suggest ways to improve, suggesting particular features that can be easily indicated during the lectures.
Thus, our main contribution to this research is:
\begin{enumerate}
\item preparing a set of didactic features feasible to annotate and detect automatically,
\item collecting a dataset with the features,
\item designing a set of deep learning models trained to determine the presence of the selected didactic feature in our dataset -- the video recordings. 
\end{enumerate}

The remainder of this work is structured as follows. Section~\ref{sec:dataset} describes our features and the collected dataset. Then, Sections~\ref{sec:experiments}, \ref{sec:text-method} and~\ref{sec:mtl-method}  introduce our experimental approach and their results. Finally, we summarize our work and indicate its main findings and limitations in~Section~\ref{sec:conclusions}.

\section{Dataset} \label{sec:dataset}



At first, we selected and analyzed teaching practices thoroughly, following the approach in~\cite{Piburn00reformedteaching,web-stp}. 
Based on these sources, we define a set of valuable didactic features from the teaching point of view, and the appropriate deep learning models can detect them. Table~\ref{tab:frequency_features} presents the designed features that are related to the main categories of Singapore Teaching Practices taxonomy~\cite{web-stp}, e.g., "activating prior knowledge," "arousing interest," "deciding on teaching aids and learning resources," "setting expectations and routines."

Once the features were defined, we chose to annotate a dataset that contains 128 lectures recorded in English (on average, one lecture lasts one and a half an hour). At least three independent research assistants annotated every lecture. Each annotates a full-length annotation of didactic features present in the lectures. Cumulatively, 380 observations were collected.
Each feature was identified as either a state or point event. State events last a specific time, i.e., they have intervals with a start and end time. Point events occur at a given moment and show the changes in behaviours or the occurrence of a feature without its specific anchoring in a given time interval. 
An event means a single didactic feature in one lecture annotated by one research assistant. 
 An observation is a set of events for one lecture annotated by one research assistant. An annotation is a set of all observations created for one lecture.

\begin{figure}[!htb]
    \centering
    \includegraphics[width=\textwidth]{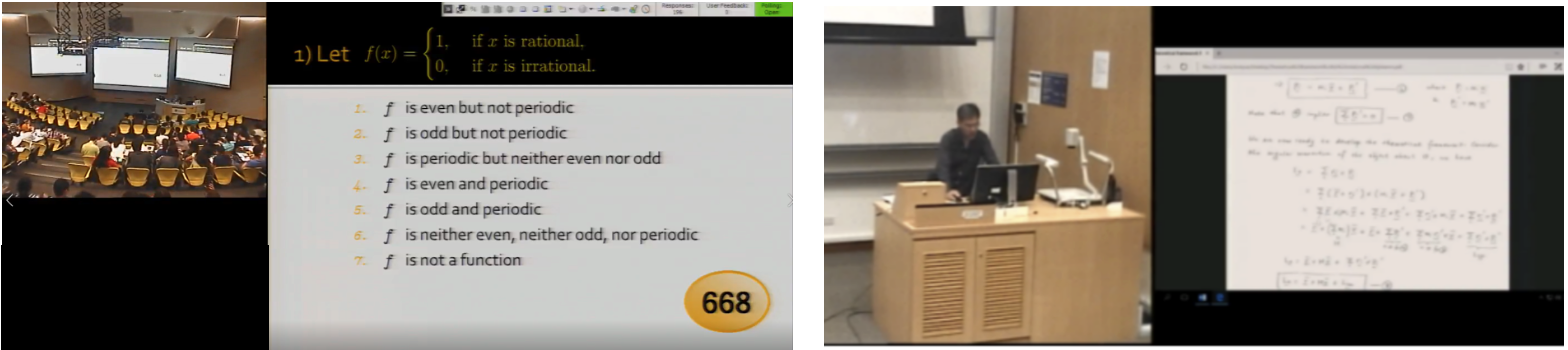}
    \caption{Example frames from the recorded lectures in our dataset}
    \label{fig:example_video}
\end{figure}

Figure~\ref{fig:example_video} presents an example of a lecture frame from our dataset. This layout often consists of two regions -- one is a camera view focused on the lecturer, and the second one is a direct input from a computer screen (often slides). 

\begin{figure}[!ht]
    \includegraphics[width=\textwidth]{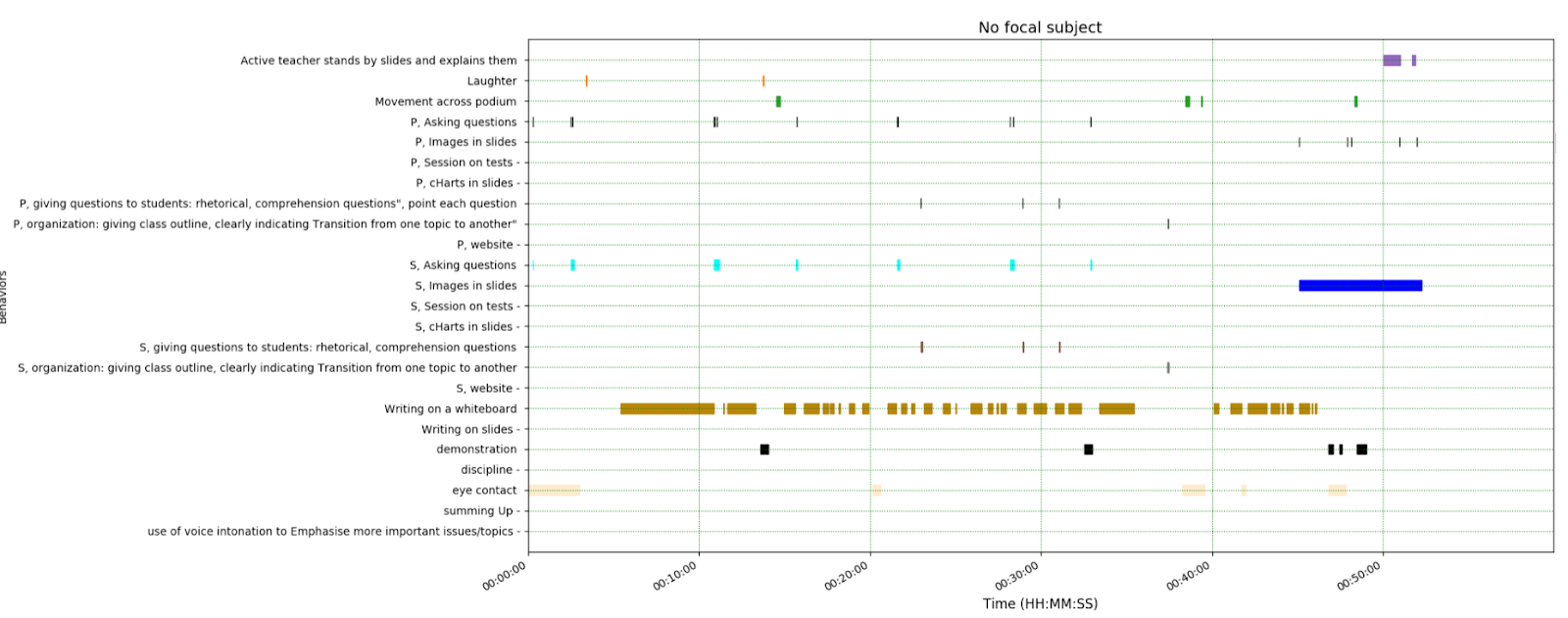}
    \caption{Example annotations, a view from BORIS programme~\cite{BORIS}}
    \label{fig:example_annotation}
\end{figure}

\section{Our Approach} \label{sec:experiments}
Our experiments concern the detection of the most frequent features in our dataset. We also choose features detected across different modalities, e.g. features that can be detected from an audio stream or its transcription, features describing slide content, or features of the teachers' behaviours (see Table~\ref{tab:frequency_features}). 
The following sections present our approach in more detail: dataset preprocessing for the deep learning tasks and their results.


\begin{table}[!ht]
\centering
\begin{tabular}{p{0.5\textwidth}p{0.15\textwidth}p{0.15\textwidth}}
Feature & Occurrences & Experiment \\  \hline
Asking questions (AQ) & 4926 & I \& I-Q \\
Giving questions, to students (GQ) & 3616 & I \& I-Q \\
Organization: giving class outline (O) & 1211 & I \\
Session on tests (S) && I \\
Active teacher stands by slides and explains them (AT) & 835 & I \\
Summing up (SU) & 72 & I \\ 
Laughter  & 315  & I-AUDIO \\
Use intonation to emphasise important issues & 200 & I-AUDIO\\
\hline
Movement across podium & 1379 & II \\
Films or animations in slides & 583 & II \\
Images in slides & 2793 & II \\
Session on tests & 854 & II \\
Charts in slides & 3356 & II \\
Website & 307 & II \\
Writing on a whiteboard & 3059 & II \\
Writing on slides & 6738 & II \\
Eye contact & 9943 & II \\ \hline
Referring to bibliography, other researchers & 55 & -- \\
Giving hints how to do something & 12 & -- \\
Students are asking questions & 151 & -- \\
Assignments & 55 & -- \\
Demonstration & 278 & -- \\
Discipline & 105 & -- \\
Students discussion & 103 & --\\
\hline
\end{tabular}
\caption{Features and number of their occurrences in our dataset, and experiments we detect them}
\label{tab:frequency_features}
\end{table}

\section{Detection of Audio-Based Features}
\label{sec:text-method}

This experiment is dedicated to the selected behavioural features based on audio stream, i.e. depending on uttered words or sounds in the lecture room. Here, we consider only audio and text-based approaches (without image data). 

Most open datasets for video annotations, especially behavioural feature recognition, are fine-grained and contain short diversified behaviours. Our goal is to detect the features that last longer during a lecture. Thus, we can use very coarse-grained portions of videos to process (e.g. windows, frames, and hop lengths), i.e., in the range of a few seconds instead of a few milliseconds, which would be required for fine-grained video annotation. It optimizes hyperparameters faster and trains models using the entire dataset, which would be too costly and require specialized hardware.

\subsection{Data Preprocessing \& Experiment Setup}
We used the Azure automatic speech recognition (ASR) service to provide data for building text models. With the service, we obtained a text with punctuation marks divided into so-called 'events.' Each event starts when the ASR engine identifies speech after a time of non-speech sound and ends before another significant gap in speech. Thus, each event consists of one or more sentences close together in time. The engine also provides timestamps of the start and stop of the whole event. We labelled each event as belonging to the given feature if its timestamps intersect with the timestamps of the feature event in our annotations.
After labelling transcription obtained from Azure, we have split the dataset into train, development (dev), and test samples in proportions 70:15:15 in such a way that all lectures of a given research assistant are in the same sample. Table~\ref{tab:ds_stats} presents statistics for our dataset. 
    \begin{table}[!htbp]
    \centering
    \begin{tabular}{l r r r}
    {} & train & dev & test \\
    \hline
    Mean length of an event (number of sentences) &  17.75 &  16.78 &  16.57 \\
    Mean duration of an event &   7.18s &   6.48s &   6.53s \\
    \hline
    Asking questions (AQ) &   8.05\% &   8.15\% &   17.89\% \\
    Giving questions to students: rhetorical... (GQ) &   7.74\% &   3.92\% &   7.69\% \\
    Organization: giving class outline...(O) &   3.58\% &   8.66\% &   3.62\% \\
    Active teacher stands by slides and explains them (AT) &   79.78\% &   32.20\% &   35.41\% \\
    Summing up (SU) &   1.21\% &   0.51\% &   0.73\% \\
    Session on tests (S) &   8.26\% &   2.41\% &   9.42\% \\
    Sum of AQ and GQ behaviors & 14.72\% & 10.45\% & 11.40\% \\
    \hline
    \end{tabular}
    \caption{\label{tab:ds_stats}Statistics of train, dev and test datasets and \% of positive samples of each feature in our dataset. Rows named after each feature contain a percentage of positive class (event -- an occurrence of the given feature)}
    \end{table}

We trained our model on two tasks: Questions only and Full task. In the Questions only task, we tried to predict whether a given sample belonged to at least one of the question features, i.e. AQ or GQ (see Table~\ref{tab:ds_stats}).
This simplified task was a  binary classification problem. We tried to predict all eight classes for audio models and six classes for text models in the Full task (see Table~\ref{tab:frequency_features}). In the text-based approach, two features ("Laughter" and "Use of voice intonation") were skipped because they express non-verbal aspects of audio, and the vital information was lost during transcription. The audio task was a more complex -- multi-label, multi-class classification problem.
    

\subsection{Detection Techniques}
\textbf{Audio networks} We employ state-of-the-art M5~\cite{very_deep_convolutional_neural_networks_for_raw_waveforms} and Wav2Letter~\cite{collobert2016wav2letter} audio networks to accomplish the feature detection on the audio stream.

\textbf{Transformers.}
We use state-of-the-art transfer learning on the uncased BERT-Base model, RoBERTa-based, and XLNet provided by HuggingFace.\footnote{https://huggingface.co/{bert-base-uncased,roberta-base,xlnet-base-cased}}
The dataset classification tasks are selected as the downstream tasks each time the training procedure and pretraining are performed on another, larger dataset. 
The networks are pre-trained on BookCorpus dataset~\cite{book_corpus}.\footnote{https://huggingface.co/datasets/bookcorpus} A feed-forward neural network head is added at the end of the Transformers and predicts classes from the dataset. 

\textbf{FastText.}
Another tool that we utilized for the simplified task of question classification is FastText~\cite{mikolov2018advances}. We applied downsampling to factor in class inequality for this model instead of using the cross-entropy loss function with weights.

\textbf{Contextual Bandits Algorithm.}
Contextual bandits algorithms\cite{jagerman2020safe},\cite{multiarmed_bandit_problem} used to perform classification were provided by Vowpal Wabbit framework.\footnote{https://vowpalwabbit.org/} 
    
\textbf{TF-IDF.}
To tokenize the data for TF-IDF, each word and punctuation mark is treated as a token. The top 10,000 most common n-grams with length $[1; 2]$ are selected and provided to the feed-forward network. 

\subsection{Results}

Audio-based detection achieves inferior results. The M5 model managed to get a non-zero F1 score, but the Wave2letter network consistently predicted the majority class. The poor performance of audio-based detection is mainly because of no open pre-trained networks and a very long training time. The audio networks' training times were about 48 hours compared to text-based networks trained within minutes (about 1 minute for FastText, Vopal Wabbit to 40 minutes for BERT, XLNet) having the same hardware settings.

The text-based models greatly outperform audio-based networks, which struggled to get any predictions correctly. The text-based networks achieve much better results despite the propagation of automatic speech recognition (ASR) transcription errors. Tables~\ref{tab:questions_only} and \ref{tab:full_task} present their results.  They are also less memory and time-consuming. Vowpal Wabbit has the best results on the questions-only task from all text-based networks. 

\begin{figure}[!htpb]
    \centering
    \includegraphics[width=0.5\textwidth]{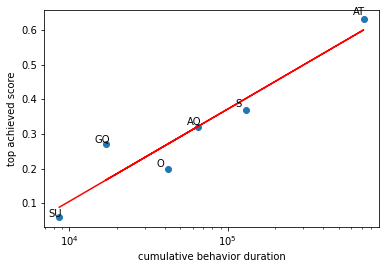}
    \caption{Correlation between cumulative duration of a behaviour and top achieved score. \behdescription}
    \label{fig:behavior_correlation}
\end{figure}

\begin{figure}[!ht]
\centering
\begin{subfigure}{.5\textwidth}
  \centering
  \includegraphics[width=.95\linewidth]{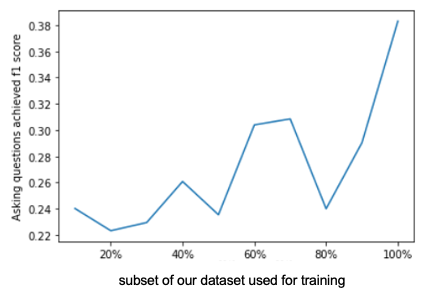}
  \caption{F1 score on our dataset}
\end{subfigure}%
\begin{subfigure}{.5\textwidth}
  \centering
  \includegraphics[width=.95\linewidth]{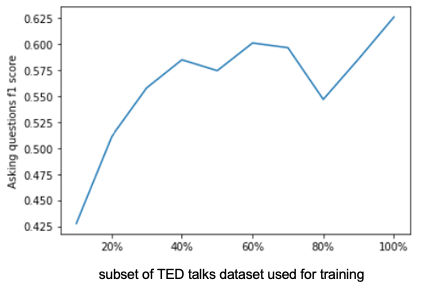}
  \caption{F1 score on TED talks dataset}
\end{subfigure}
\caption{F1 score achieved on a simplified \textit{Asking questions} task using BERT on a subset of available data in our datasets and similarly prepared dataset from TED talks}
\label{fig:subset_results}
\end{figure}

We can also notice that the prediction of different features varies greatly. The difference roughly corresponds to the number of occurrences in our dataset. The three features with the highest number of occurrences and the most extended duration times are also the ones with the highest F1 score (see Table~\ref{tab:full_task} and Figure~\ref{fig:behavior_correlation}). 
Figure~\ref{fig:subset_results} presents F1 scores when training is performed on BERT using only a subset of available data from 10\% to 100\% (here, we compared the results of our dataset with a similarly prepared TED talks dataset). 

\begin{table}[!htbp]
\centering
\begin{tabular}{lllll}
\textbf{Model} & \textbf{Accuracy} & \textbf{Precision}
& \textbf{Recall} & \textbf{F1} \\
\hline
BERT & 0.831 & 0.332 & 0.453 & 0.383 \\
VowpalWabbit & 0.757 & 0.481 & 0.387 & \textbf{0.429} \\
FastText & 0.745 & 0.442 & 0.321 & 0.373 \\
RoBERTa & 0.116 & 0.116 & \textbf{1.00} & 0.207 \\
XLNet & 0.690 & 0.177 & 0.459 & 0.255 \\
TF-IDF & 0.884 & 0.461 & 0.100 & 0.164 \\
\end{tabular}
\caption{Results for the "Asking questions" (AQ \& GQ) prediction task}
\label{tab:questions_only}
\end{table}
\begin{table}[!ht]
    \centering
    \begin{tabular}{lllllll}
    \textbf{Model} & \textbf{AQ} & \textbf{GQ} & \textbf{O} & \textbf{S} & \textbf{AT} & \textbf{SU} \\ \hline
    BERT & 0.29 & \textbf{0.27} & 0.03 & 0.08 & 0.31 & 0.02 \\
    VowpalWabbit & \textbf{0.32} & \textbf{0.27} & 0.17 & \textbf{0.37} & 0.52 & 0.02 \\
    FastText & 0.04 & 0.03 & 0.03 & 0.16 & 0.43 & 0.0 \\
    RoBERTa & 0.28 & 0.25 & \textbf{0.20} & 0.20 & 0.45 & \textbf{0.06} \\
    XLNet & 0.29 & 0.24 & 0.19 & 0.21 & 0.45 & 0.02 \\
    TF-IDF & 0.08 & 0.09 & 0.05 & 0.21 & \textbf{0.63} & 0.0 \\
    \end{tabular}
    \centering
    \caption{\label{tab:full_task}F1 scores on Full task -- all the text features.\behdescription}
\end{table}

Training and hyperparameter optimization are much easier on a purely text-based dataset due to: (1) much smaller size (the datasets are lowered no less than 25,000-fold in size), (2) greater speed, and simplification of the task (the model does not have to extract data from raw audio or audio features). Results vary greatly between the features. The features with lower count (e.g. \textit{Summing up}) have much lower F1 score than those with higher observation count (e.g. \textit{Asking questions}, \textit{Giving questions to students}).
It is essential how text samples are prepared, i.e., a balance must be structuralized between making not too long sentences and providing enough context for correct prediction.

The annotations seem to be less reliable than expected; for example, out of 6,500 sentences containing question marks, only 2,500 are labelled as questions. Sentences with question marks that are not automatically classified as questions are mostly: questions seeking affirmations ("right?", "um?") or questions that are part of explanations ("So, how should we approach this problem? Let's try this way."). Moreover, these sentences are not indicated as questions by annotators yet are rhetorical questions which means they belong to the feature \textit{giving questions to students: rhetorical questions}.

Figure \ref{fig:subset_results} and \ref{fig:behavior_correlation} suggest that results can be significantly improved by increasing the size of our datasets and the consistency of annotations, even in comparison to a simulated dataset with TED talks, similarly preprocessed. 


\section{Multi-Task Learning for Visual Features Detection}\label{sec:mtl-method}

In this experiment, we employed multi-task learning (MTL) \cite{mtl} for detecting visual features, such as "chart in slides" or "movement across the podium." They appear primarily in the view of the lecturer and presentation screen (see Experiment II in Table~\ref{tab:frequency_features}). This experiment also analyses whether features or their groups can be recognized in one model simultaneously with multi-task learning.

We omit the "active teacher stands by slides and explains them" feature because the essential part is the explanation part, which is what the lecturer is saying. On the other hand, we have selected "session on tests" because these sessions are test sessions with questions appearing on slides. 

\subsection{Data Preprocessing}
Here, the prepared dataset consists of selected frames from videos from our dataset. Because of technical limitations, we propose an algorithm that makes predictions independently for every frame. We chose a frame in the middle of every occurrence of the feature (an event recorded in our dataset). It creates a dataset of less correlated events, capturing exactly one observation per recorded event.

\begin{figure}[!ht]
    \centering
    \includegraphics[width=\textwidth]{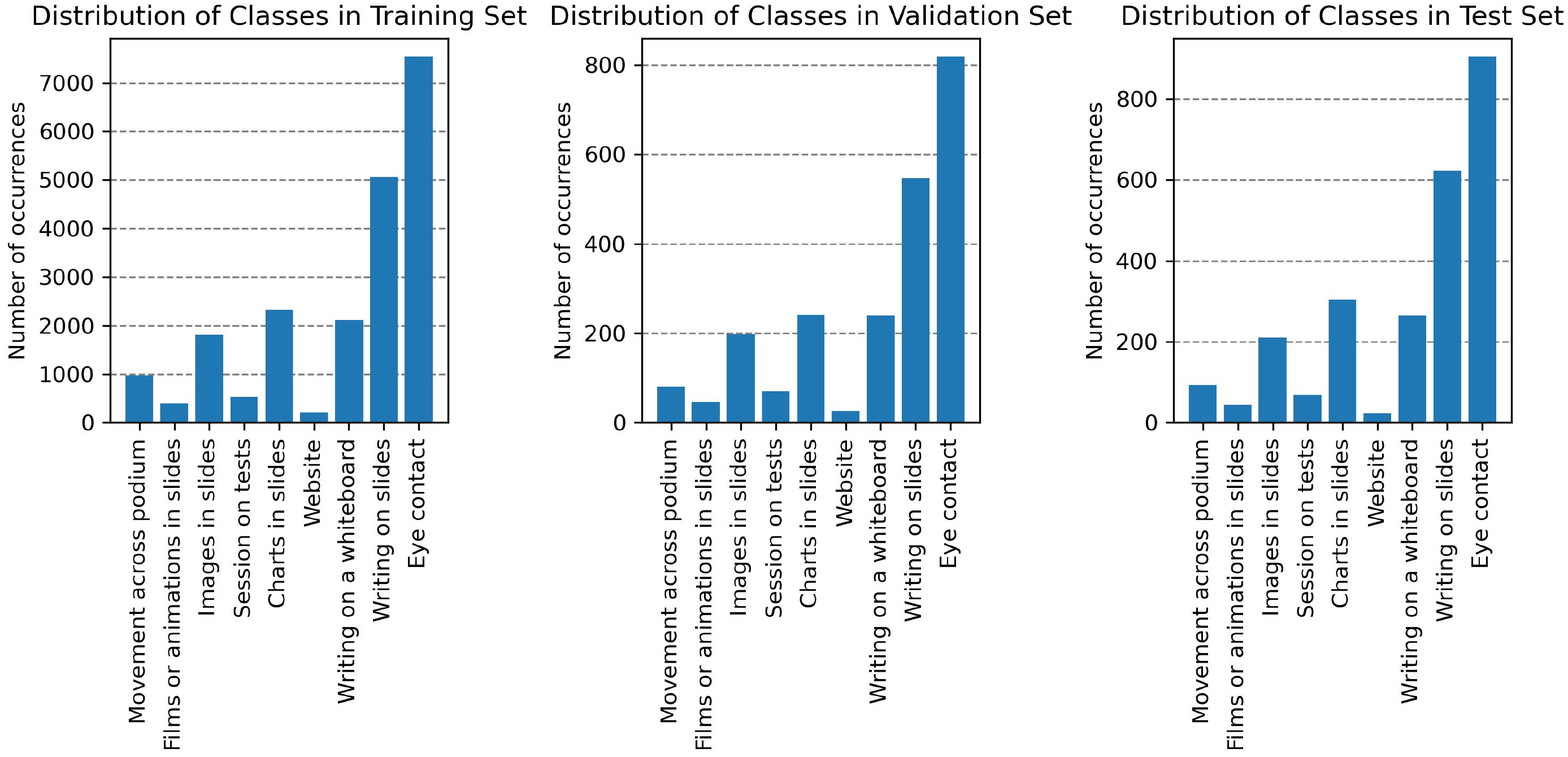}
    \caption{Distribution of the features in Experiment II (MTL) 
    }
    \label{fig:FeatureRecognitionModel:distributions_of_classes}
\end{figure}

We had roughly 40,000 events recorded. After removing duplicates, events happening simultaneously, and filtering only events of interest relevant to visual data, we were left with 25,764. Our dataset consists of two primary sources of resemblances in data: frames in one video are similar. The dataset comprises several series of lectures that typically look similar between videos. We split our data into three independent sets based on the whole series of lectures for training purposes. The test set consists of 2,203 events, and the validation set consists of 2,534, which leaves 21,027 events for the training dataset. 
Distributions of features in the training, testing and validation datasets are similar (see Figure~\ref{fig:FeatureRecognitionModel:distributions_of_classes}). 

\subsection{Network Architecture \& Training Setup}
\begin{figure}[!ht]
    \centering
    \includegraphics[width=0.8\textwidth]{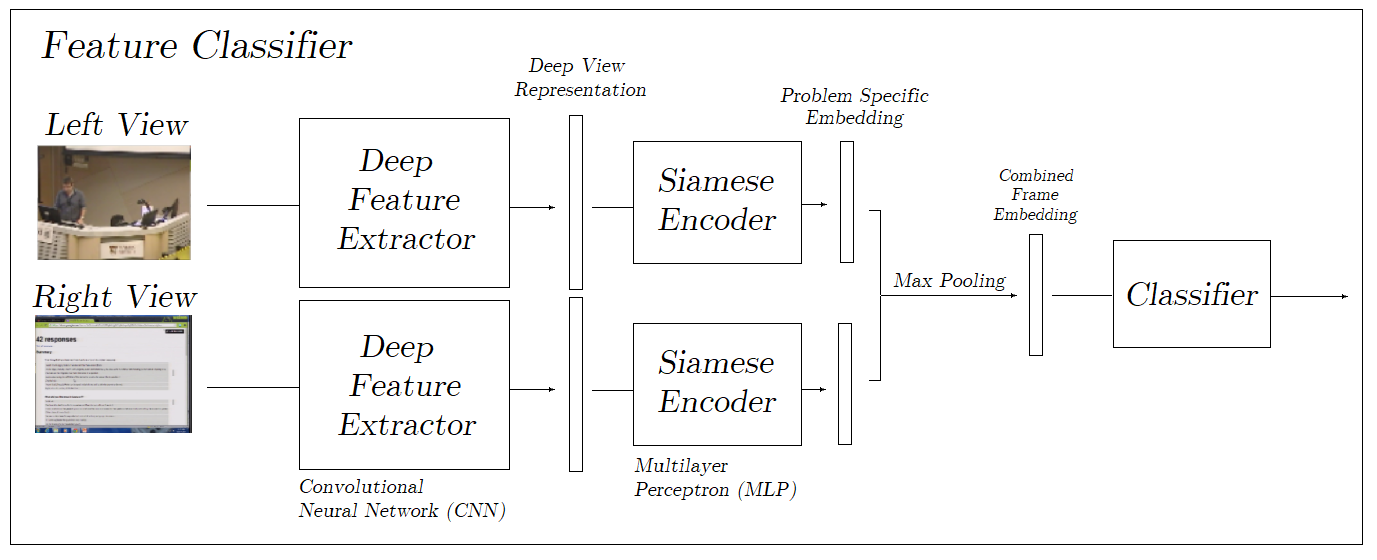}
    \caption{Feature detection model architecture outline for MTL}
    \label{fig:FeatureRecognitionModel:feature_classifier_outline}
\end{figure}

The designed solution consists of feature extraction with a pre-trained model, a siamese encoder, and a classifier. Figure~\ref{fig:FeatureRecognitionModel:feature_classifier_outline} presents the architecture scheme. After splitting the video frame into separated information channels, each view is embedded by a feature extractor. Then deep view representations go into the siamese encoder, a neural network that extracts relevant features for our problem. In the end, view representations are combined with max-pooling and are passed to the classifier.
This classifier -- here a simple multi-layer perceptron -- returns class scores for the nine classes (each responsible for detecting one feature) and thus makes the final prediction. 

As a pre-trained network for the feature extractor in our architecture, we examined three basic architectures: AlexNet \cite{AlexNet}, VGG19 \cite{VGG} and ResNet50 \cite{ResNet}.  Two of them are tested twice for the last layer representation and the third layer of the last intermediate representation. 
Finally, we trained five networks, one for each representation, and tested each model applying cross-entropy loss \cite{Goodfellow} and second type -- cross-entropy loss with weighting the results. The training is repeated at least five times for every training setting. 

\subsection{Results}
\begin{figure}[!htb]
\begin{subfigure}{\textwidth}
    \centering
    \includegraphics[width=\textwidth]{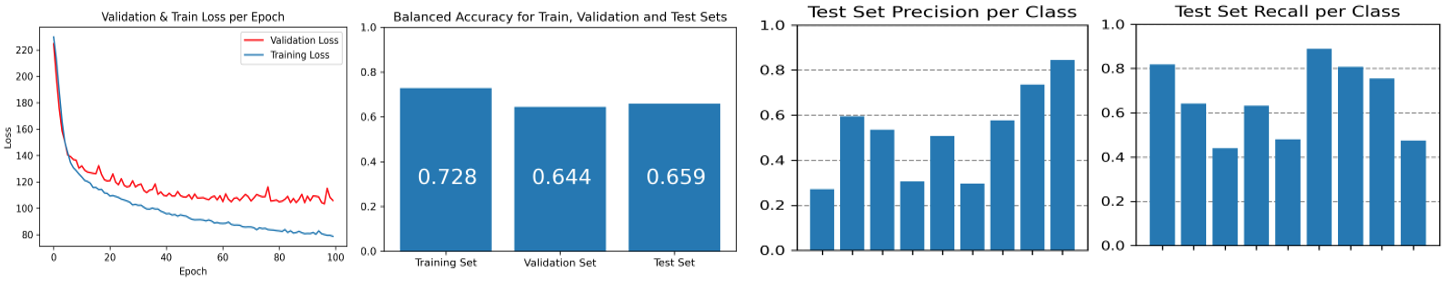}
    \caption{AlexNet with last representation layer as a feature extractor}
\end{subfigure}%
\\
\begin{subfigure}{\textwidth}
  \centering
  \includegraphics[width=\textwidth]{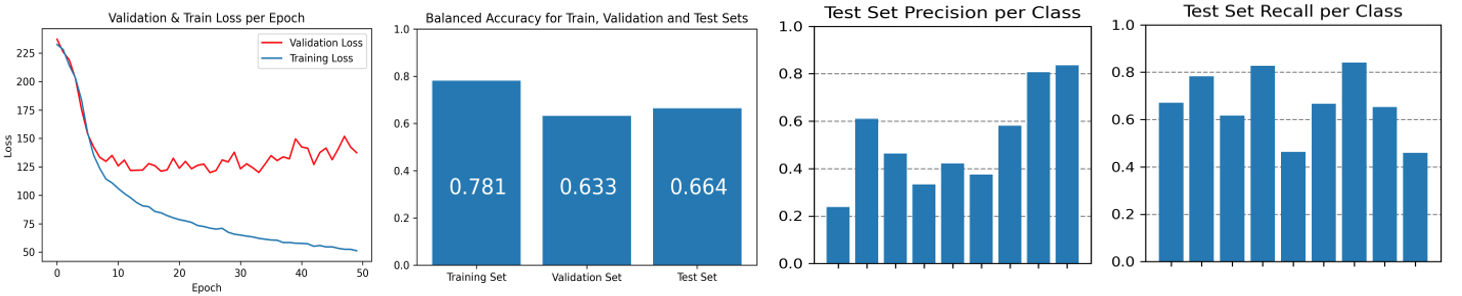}
    \caption{AlexNet with third representation layer as a feature extractor}
 \end{subfigure}
    \caption{Results for our networks with AlexNet as a feature extractor with balanced loss while training. From the left to the right: network learning curve, balanced accuracies, precision and recall on the test set for each class -- the detected features, i.e. in the given order: "Movement across podium", "Films or animations in slides", "Images in slides", "Session on tests", "Charts in slides", "Websites", "Writing on a whiteboard", "Writing on slides", "Eye contact"}
    \label{fig:mtl-best-model}
\end{figure}

Networks performance appears to be much better than random overall (we have nine features to recognize). Training with imbalanced loss, the AlexNet with the third representation layer as a feature extractor achieves above 70\% accuracy for the test. However, our dataset is imbalanced, so a more reliable measure is balanced accuracy, an average recall score of each class/feature. The best overall balanced accuracy was about 66\% achieved by the AlexNet network, also trained on its last third hidden representation with an imbalanced loss function (see Figure~\ref{fig:mtl-best-model}). For balanced loss, the last intermediate representation of the VGG19 achieves similar results to the AlexNet. 

Another vital thing our analysis shows is that classes might fight. 
First, the distribution of network prediction trained on balanced loss appears to be more uniform. Those networks do not favour the most frequent classes and guess much more often than a less frequent feature is visible in the video lecture.

There are differences in how networks make mistakes. They commonly confuse "images in slides" class with "charts in slides". Another common mistake is confusing "movement across podium" with "eye contact", especially for networks trained with a balanced loss function. It is understandable, as these feature pairs rely on similar data and often co-occur.




Multi-task learning has great theoretical potential. However, we argue that a more specialized approach is needed for our problem, but at the same time, the use of MTL \cite{mtl} in future work may still be beneficial. The MTL approach brings the possibility of learning distinctions between even features. It can be far more efficient than other deep learning tasks since characteristics can be shared between different features and particular occurrences. Given our results that are not yet acceptable for professional use, they are not very bad. It is still possible that this approach can work if supplied with a better feature extractor and trained on a more extensive dataset.




\section{Conclusions}\label{sec:conclusions}

Our contribution in this work was defining features that can give tips to assess qualitatively didactic aspects of lectures. Then we collected and annotated a dataset that allowed training and testing of deep learning networks for automatic detection and recognition of these features.
Our results showed the usefulness of our work. We delivered an approach to prepare and preprocess the data of video recording. Also, we showed the main directions for deep learning methods. Especially we bypassed audio processing and modelling because it needed extensive computing performance, much more than text processing.

However, the dataset should be more extensive and diversified to achieve stable results. Also, network architectures should be more sophisticated and tuned. 
Also, our results showed an urge for more effort to get proper annotations, such as: defining extended annotation protocol, employing more annotators, and training them to be more consistent.



%
%
\section*{Acknowledgements}
{\small 
The research was funded by the Centre for Priority Research Area Artificial Intelligence and Robotics of Warsaw University of Technology within the Excellence Initiative: Research University (IDUB) programme (grant no 1820/27/Z01/POB2/2021) and by the RENOIR project under the EU program Horizon 2020 under project contract no 691152.\\
We want to thank Sylwia Sysko-Romanczuk for her ideas and help in the structuralization of this research.}

\bibliographystyle{unsrtnat}
\bibliography{references}  

\end{document}